\documentclass[a4paper,fleqn]{cas-dc}

\usepackage[numbers]{natbib}
\usepackage{graphicx}%
\usepackage{multirow}%
\usepackage{amsmath,amssymb,amsfonts}%
\usepackage{amsthm}%
\usepackage{mathrsfs}%
\usepackage{tabularx}
\usepackage[linesnumbered, ruled, vlined]{algorithm2e}
\usepackage[title]{appendix}%
\usepackage{xcolor}%
\usepackage{textcomp}%
\usepackage{manyfoot}%
\usepackage{booktabs}%
\usepackage{makecell} 
\usepackage{listings}
\usepackage{graphicx}
\usepackage{color}
\usepackage{tabularx}
\usepackage{tabulary}
\usepackage{longtable}
\usepackage{booktabs} 
\usepackage{siunitx} 
\usepackage{caption}
\usepackage{booktabs}
\usepackage{siunitx}
\usepackage{makecell}
\usepackage{subcaption}
\usepackage{url}
\usepackage{geometry}
\usepackage{fancyhdr}

\begin{document}

\title [mode = title]{MRD-LiNet: A Novel Lightweight Hybrid CNN with Gradient-Guided Unlearning for Improved Drought Stress Identification}   



\author[1,2]{Aswini Kumar Patra}
\ead{aswinipatra@gmail.com/akp@nerist.ac.in}

\credit{Conceptualization of this study, Methodology, Writing, Coding and Implementation}

\affiliation[1]{organization={Department of Computer Science and Engineering, North Eastern Regional Institute of Science and Technology},
                addressline={Nirjuli}, 
                city={Itanagar},
                state={Arunachal Pradesh},
                postcode={791109},
                country={India},
                }

\author[2]{Lingaraj Sahoo}
\cormark[1]
\ead{ls@iitg.ac.in}

\affiliation[2]{organization={Depart ment of Bioscience and Bio-Engineering, Indian Institute of Technology Guwahati},
                addressline={North Guwahati},  
                state={Assam},
                 postcode={781039},
                country={India},
               }

\cortext[cor1]{Corresponding author}

\begin{abstract}
Drought stress is a major threat to global crop productivity, making its early and precise detection essential for sustainable agricultural management. Traditional approaches, though useful, are often time-consuming and labor-intensive, which has motivated the adoption of deep learning methods. In recent years, Convolutional Neural Network (CNN) and Vision Transformer architectures have been widely explored for drought stress identification; however, these models generally rely on a large number of trainable parameters, restricting their use in resource-limited and real-time agricultural settings.
To address this challenge, we propose a novel lightweight hybrid CNN framework inspired by ResNet, DenseNet, and MobileNet architectures. The framework achieves a remarkable 15-fold reduction in trainable parameters compared to conventional CNN and Vision Transformer models, while maintaining competitive accuracy. In addition, we introduce a machine unlearning mechanism based on a gradient norm-based influence function, which enables targeted removal of specific training data influence, thereby improving model adaptability.
The method was evaluated on an aerial image dataset of potato fields with expert-annotated healthy and drought-stressed regions. Experimental results show that our framework achieves high accuracy while substantially lowering computational costs. These findings highlight its potential as a practical, scalable, and adaptive solution for drought stress monitoring in precision agriculture, particularly under resource-constrained conditions.
\end{abstract}



\maketitle

%
\section{Introduction}
Drought stress is one of the most severe abiotic factors threatening global crop productivity and food security. Its early and precise identification is essential for timely intervention, efficient resource management, and sustaining agricultural yields~\cite{araus_field_2014,2021_Ali,farooq_plant_2009}. Conventional assessment methods, including manual field inspections and physiological measurements, are often labor-intensive, subjective, and not scalable for large-scale agricultural operations. Recent advances in deep learning, particularly Convolutional Neural Networks (CNNs), have enabled automated and high-throughput analysis of plant health from aerial imagery, opening new possibilities for precision agriculture~\cite{kamilaris_review_2018}.

Initial research on drought and water stress relied heavily on traditional machine learning and handcrafted features. For example, Zhuang et al.~\cite{zhuang_early_2017} used color and texture features with a gradient boosting decision tree (GBDT) for water stress detection in maize, though later studies showed CNNs significantly outperformed GBDT in both accuracy and robustness~\cite{an_identification_2019}. Further progress has been made through advanced imaging modalities and machine learning pipelines, such as hyperspectral imaging in groundnut~\cite{sankararao_machine_2023}, transfer learning with DenseNet-121 for soybean drought severity~\cite{ramos-giraldo_drought_2020}, and hybrid CNN-LSTM approaches for chickpea~\cite{azimi_intelligent_2021}. Comparative studies have also reported GoogLeNet as highly accurate for multiple crops~\cite{chandel_identifying_2021} and tree-based classifiers (e.g., Random Forest, Extra Trees) as effective for chlorophyll fluorescence-based stress detection~\cite{gupta_drought_2023}.  

In potato crops, Butte et al.~\cite{butte_potato_2021} applied deep learning to multi-modal aerial imagery, while Patra et al.~\cite{patra2024explainable} proposed an explainable CNN framework that improved classification accuracy. Other works confirmed the reliability of SVMs and Random Forests with hyperspectral data~\cite{chen_hyperspectral_2022,dao_plant_2021}, while Goyal et al.~\cite{goyal_deep_2024} designed a custom CNN surpassing state-of-the-art models for maize drought detection.

Recently, Vision Transformers (ViTs) have gained traction in plant stress identification due to their self-attention mechanism, which captures long-range dependencies and global context~\cite{dosovitskiy_image_2021,chen_transunet_2021,liu_swin_2021}. ViTs and hybrid variants have achieved superior performance in plant disease and stress classification~\cite{bhowmik_customised_2024,singh_effective_2024,borhani_deep_2022,thakur_vision_2023}. For example, ViTs have outperformed Inception V3 in tomato disease classification~\cite{barman_vit-smartagri_2024}, while Parez et al.~\cite{parez_visual_2023} demonstrated accuracy gains with a fine-tuned ViT using fewer parameters. Recent innovations include reduced transformer encoders for drought stress identification~\cite{patra2025explainable}, lightweight modules~\cite{gole_trincnet_2023}, attention-head optimization~\cite{thai_formerleaf_2023}, and transformer–CNN hybrids~\cite{vallabhajosyula_novel_2024}.  
Despite these successes, most works emphasize accuracy, with limited focus on parameter efficiency—a key barrier for real-time deployment in resource-constrained agricultural environments.

Both advanced CNNs and ViT-based models remain computationally demanding and parameter-heavy, which limits their suitability for edge deployment and large-scale agricultural applications. To address this challenge, some recent attempts have explored lightweight models for drought stress detection, such as the approaches proposed by Patra et al.~\cite{patra2024explainable,patra2025explainable}, Li et al. \cite{li_pmvt_2023} and Gole et al.~\cite{gole_trincnet_2023}. As CNN architectures such as ResNet~\cite{koonce2021resnet}, DenseNet~\cite{iandola2014densenet}, and MobileNet~\cite{howard2017mobilenets} employ relatively fewer parameters while incorporating architectural innovations that enhance both accuracy and efficiency, their combined strengths present a promising direction. Therefore, developing a lightweight hybrid model that integrates these architectures can be highly effective and is the focus of investigation in this work.

Equally important, deployed model must adapt to evolving environmental conditions, mislabeled data, and outliers. In this context, \textit{machine unlearning}—the process of selectively removing the influence of specific training data—has emerged as a tool for enhancing adaptability, and error correction ~\cite{xu2024machine, li2025overview}. While the bulk of unlearning research focuses on privacy, there are emerging connections to plant stress detection by supporting adaptive models. Bourtoule et al.\cite{bourtoule2021machine} proposed the foundational “Sharded, Isolated, Sliced, and Aggregated” (SISA) approach for efficient unlearning in deep models, while Cao and Yang \cite{cao2015towards} earlier formalized the notion of statistical unlearning by bounding the effect of deleted samples. Later, Ginart et al.\cite{graves2021amnesiac} explored “amnesiac” machine learning that selectively forgets training data. Practical advances in deep neural networks include gradient-based influence estimation for fast unlearning~\cite{golatkar2020eternal}, certified removal guarantees~\cite{guo2019certified}, and unlearning mechanisms in computer vision tasks~\cite{kurmanji2023towards}. These studies provide a strong foundation for adapting unlearning to agricultural stress identification, where mislabeled samples and environmental noise are common.

In this work, we propose a framework that addresses both efficiency and adaptability. Specifically, we:  
\begin{itemize}
    \item Design a novel lightweight hybrid CNN architecture, inspired by ResNet, DenseNet, and MobileNet, that achieves a 15-fold reduction in trainable parameters compared to CNN and ViT models, while maintaining high accuracy for drought stress identification.  
    \item Develop a gradient norm-based machine unlearning mechanism that enables selective removal of the influence of specific training samples, thereby enhancing adaptability.  
    \item Validate the proposed framework on a challenging real-world aerial imagery dataset of potato crops~\cite{butte_potato_2021}, demonstrating superior efficiency, performance, and scalability compared to state-of-the-art methods.  
\end{itemize}

\section{Material and Methods}
\subsection{Data Set Description}
The aerial imagery of potato crops employed in this study was obtained from a publicly available, multi-modal dataset \cite{potato_data, butte_potato_2021}. The dataset was collected at the Aberdeen Research and Extension Center, University of Idaho, and is specifically designed to support machine learning research in crop health monitoring for precision agriculture. Image acquisition was performed using a Parrot Sequoia multi-spectral camera mounted on a 3DR Solo drone. This setup provided high-resolution RGB images at $4,608 \times 3,456$ pixels, along with four monochrome channels—green (550nm), red (660nm), red-edge (735nm), and near-infrared (790nm)—each with a resolution of $1,280 \times 960$ pixels. The drone operated at a low altitude of 3 meters to capture detailed visual information, with a focus on identifying drought stress in Russet Burbank potato plants affected by premature senescence.

For this work, the dataset includes 360 RGB image patches in JPG format, each sized 750×750 pixels. These patches were generated from the original high-resolution aerial images through cropping, rotation, and resizing. To expand the training data, data augmentation techniques were applied to 300 of these images following the method described by Butte et al. \cite{butte_potato_2021}, resulting in a total of 1,500 training images. The remaining 60 images were set aside as a test set, with no augmentation applied, to ensure an unbiased evaluation of model performance.

Accurate classification required labeled data with annotated regions of interest. In this study, regions containing healthy and drought-stressed potato plants were manually annotated using the LabelImg tool \cite{noauthor_labelimg_nodate}. Healthy plants were visually identified by their green coloration, while stressed plants appeared yellowish. Bounding boxes were drawn around these regions, and the corresponding class labels and coordinates were saved to generate ground truth annotations for model training.

In addition to RGB images, the dataset also provides corresponding spectral patches (red, green, red-edge, and near-infrared), each measuring 416×416 pixels. However, due to the limited resolution of these monochrome images, only the RGB channels were utilized in this study.

\begin{figure*}[h!]
    \centering
    \begin{subfigure}{0.45\textwidth}
        \centering
        \includegraphics[width=0.9\textwidth]{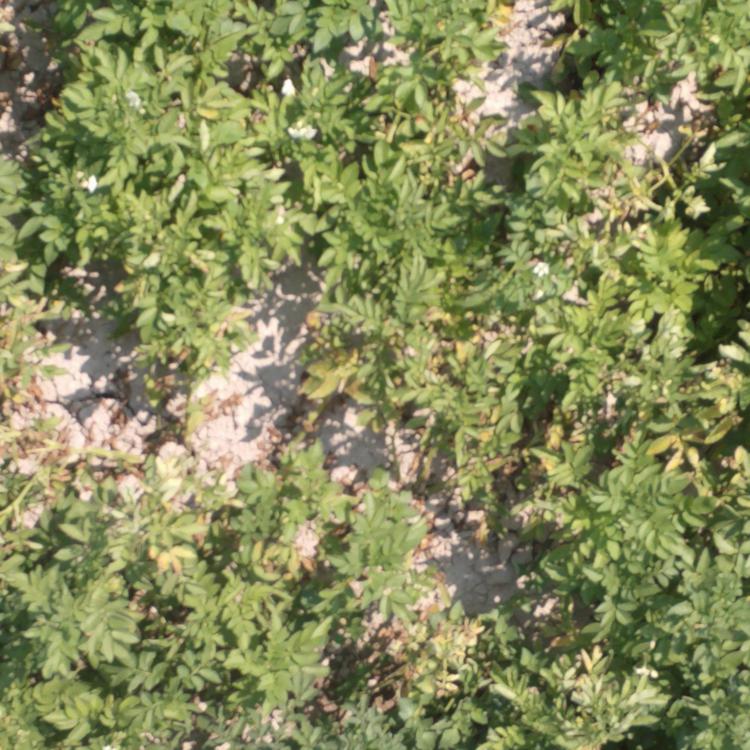}
        \caption{}
        \label{fig:rgb_sa}
    \end{subfigure}
    \begin{subfigure}{0.45\textwidth}
        \centering
        \includegraphics[width=0.9\textwidth]{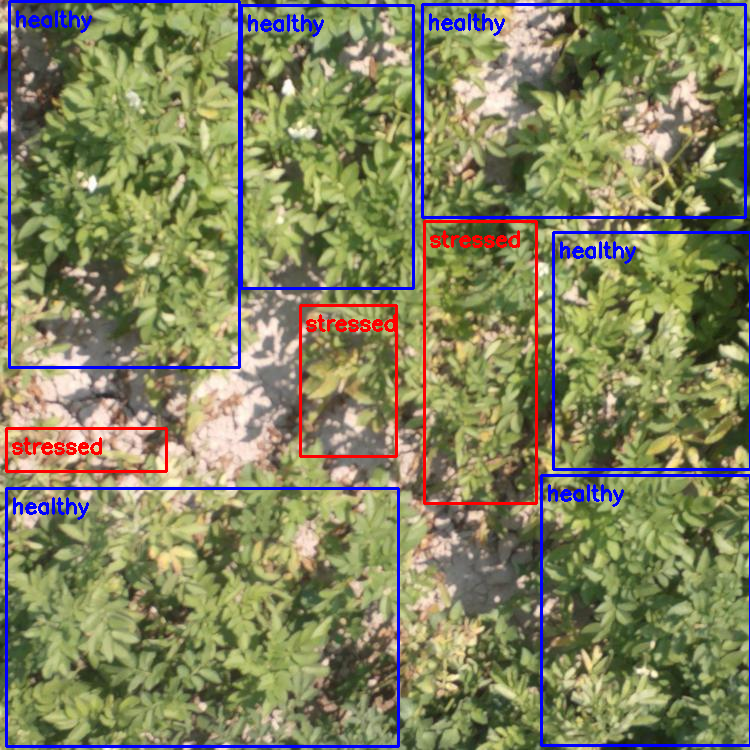}
        \caption{}
        \label{fig:hea_st}
    \end{subfigure}
    \caption{Field images showing \subref{fig:rgb_sa}) Sample RGB image and \subref{fig:hea_st}) Healthy and Stressed Labels.}
    \label{fig:sample_rgb}
\end{figure*}

The final augmented training set consisted of 1,500 images, while the test set comprised 60 unique images. Annotated windows (rectangular patches) were extracted from both training and test images based on the bounding box information, with each window labeled as either “healthy” or “stressed.” As depicted in Fig. \ref{fig:sample_rgb}, Fig. \ref{fig:rgb_sa} shows an original RGB image, whereas Fig. \ref{fig:hea_st} illustrates the extracted healthy and stressed regions. In the provided example, six windows correspond to healthy areas and three to stressed ones. After extraction, the final training dataset contained 11,915 stressed and 8,200 healthy image patches. The test set contains 734 stressed and 401 healthy windows extracted from the 60 test images.

\section{Methodology}
The proposed model is a novel convolutional neural network (CNN) architecture that synergistically combines the principles of \textbf{MobileNetV2}, \textbf{Resnet} and \textbf{DenseNet}. The design goal is to achieve an optimal balance between model \textit{efficiency}, \textit{representational power}, and \textit{gradient flow}, critical for binary image classification tasks where both precision and computational feasibility are essential. 

\begin{figure*}[!htb]
        \centering
		\includegraphics[scale=1.3]{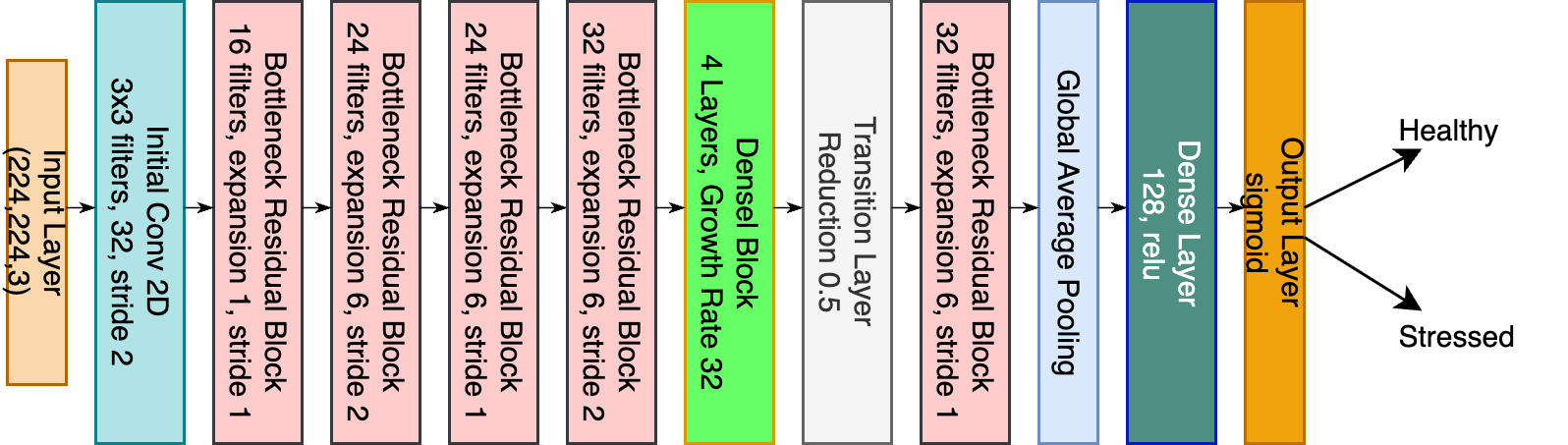}
		\caption{Schematic diagram of the lightweight network architecture} \label{frame_ul1}
\end{figure*}

The proposed framework begins with an initial convolutional layer for low-level feature extraction, followed by four residual blocks that enable deeper feature learning through skip connections. A dense block is then employed to encourage feature reuse and efficient representation learning, after which a transition layer reduces dimensionality and controls model complexity. This is followed by a bottleneck block that captures the most salient features in a compact form. The high-level representations are aggregated using global average pooling, passed through a fully connected dense layer for further processing, and finally mapped to the output layer for prediction. The overall architecture of the proposed framework is depicted in Fig. \ref{frame_ul1}, illustrating the sequence of layers from the initial convolution through to the output layer. The detailed steps and parameters of each constituent block in the proposed model are summarized in Algorithm \ref{alg:light_weight_hybrid}. A brief description of each block follows. 

\subsubsection{Input and Initial Convolution}

The model processes RGB input images with a resolution of $224 \times 224$. An initial convolutional layer applies 32 filters of size $3 \times 3$ with stride 2 and ``same'' padding. This layer is followed by batch normalization and ReLU6 activation to standardize feature distributions and introduce non-linearity while preserving numerical stability for low-precision computation.

\subsubsection{Bottleneck Residual Blocks}

Inspired by MobileNetV2, the Bottleneck Residual Block comprises three phases:

\begin{itemize}
    \item \textbf{Expansion Phase}: A $1 \times 1$ pointwise convolution increases channel dimensionality, enabling a richer representation of features.
    \item \textbf{Depthwise Convolution}: A $3 \times 3$ depthwise convolution performs lightweight spatial filtering independently across channels, drastically reducing computational cost.
    \item \textbf{Projection Phase}: A second $1 \times 1$ convolution projects the features back to a lower dimension.
\end{itemize}

These blocks integrate \textit{skip connections} when input and output dimensions match, preserving information and enhancing gradient flow during backpropagation. Their significance lies in reducing computation while retaining accuracy—ideal for scalable deep learning models.

\subsubsection{Dense Block}

A Dense Block consisting of four convolutional units is embedded mid-network. Each unit applies BatchNorm $\rightarrow$ ReLU $\rightarrow$ $3 \times 3$ Conv2D and concatenates its output with all preceding feature maps. This design enforces: feature reuse across layers, improved gradient propagation, diminished vanishing gradient issues, especially in deeper networks.

Dense connectivity introduces a collective memory mechanism, where each layer benefits from the cumulative knowledge of all previous layers, enhancing both convergence speed and generalization.

\subsubsection{Transition Layer}

To manage the increasing number of channels from the Dense Block, a Transition Layer is applied. It performs:

\begin{itemize}
    \item Channel compression using a $1 \times 1$ convolution,
    \item Downsampling via $2 \times 2$ average pooling.
\end{itemize}

This module plays a \textit{regularization role}, reducing model complexity and the risk of overfitting, while maintaining vital feature information. The transition also helps in controlling memory usage and computation.

\subsubsection{Final Processing and Classification}

Following the transition, a final Bottleneck Residual Block is applied. The feature map is then reduced to a 1D vector using Global Average Pooling (GAP), which reduces overfitting and parameter count compared to fully connected layers. A Dense Layer with 128 units and ReLU activation processes the features, followed by a sigmoid-activated output unit to generate binary classification results.

\subsubsection{Optimization Strategy}

The model is trained using the \textit{Adam} optimizer with a scheduled exponential learning rate decay, which gradually lowers the learning rate to promote stable convergence. The initial learning rate is set to 0.001, with a decay rate of 0.9 applied every \(2 \times \text{steps\_per\_epoch}\). Here, the steps per epoch are calculated as  
\[
\text{steps\_per\_epoch} = \left\lfloor \frac{\text{num\_train\_samples}}{\text{batch\_size}} \right\rfloor.
\]

The \textbf{binary crossentropy} loss function is employed to handle binary classification, particularly in cases of class imbalance. The optimizer’s learning rate schedule is designed to balance fast convergence at the start with fine-grained updates toward the end.

\begin{algorithm*}
\caption{Customized lightweight CNN Model}
\label{alg:light_weight_hybrid}
\KwIn{Input shape: $224 \times 224 \times 3$, Training samples: $num\_train\_samples$}
\KwOut{Trained CNN model with bottleneck residual blocks and DenseNet layers}

\SetKwFunction{BottleneckBlock}{BottleneckBlock}
\SetKwFunction{DenseBlock}{DenseBlock}
\SetKwFunction{TransitionLayer}{TransitionLayer}

\textbf{Sub-functions:}
\begin{itemize}
    \item \BottleneckBlock{$x$, $filters$, $expansion\_factor$, $stride$}:
    \begin{itemize}
        \item Conv2D($filters \times expansion\_factor$, $(1, 1)$), BatchNorm, ReLU(6.0)\;
        \item DepthwiseConv2D$(3, 3)$, stride = $stride$, BatchNorm, ReLU(6.0)\;
        \item Conv2D($filters$, $(1, 1)$), BatchNorm\;
        \item If $stride == 1$ \& $x.shape[-1] == filters$, skip connection\;
    \end{itemize}

    \item \DenseBlock{$x$, $num\_layers$, $growth\_rate$}:
    \For{$i = 1$ to $num\_layers$}{
    \begin{itemize}
        \item BatchNorm, ReLU, Conv2D($growth\_rate$, $(3, 3)$)\;
        \item Concatenate input $x$ with Conv2D output\;
    \end{itemize}    
    }
    
    \item \TransitionLayer{$x$, $reduction$}:
    \begin{itemize}
        \item BatchNorm, ReLU, Conv2D($reduction \times x.shape[-1]$, $(1, 1)$)\;
        \item AveragePooling $(2, 2)$, strides = 2\;
    \end{itemize}
\end{itemize}

\textbf{Main Model:}
\begin{itemize}
    \item Input: $224 \times 224 \times 3$\;
    \item Conv2D(36, $(3, 3)$), stride = 2, BatchNorm, ReLU(6.0)\;
    
    \item \BottleneckBlock{$x$, 16, 1, 1}, \BottleneckBlock{$x$, 24, 6, 2}\;
    \item \BottleneckBlock{$x$, 24, 6, 1}, \BottleneckBlock{$x$, 32, 6, 2}\;

    \item \DenseBlock{$x$, 4, 32}\;
    \item \TransitionLayer{$x$, 0.5}\;

    \item \BottleneckBlock{$x$, 32, 6, 1}\;

    \item Global Average Pooling\;
    \item Dense(128, ReLU), Dense(1, Sigmoid)\;

    \item Initial learning rate: 0.001\;
    \item Steps per epoch: $num\_train\_samples / batch\_size$\;
    \item Exponential decay learning rate schedule\;

    \item Compile model (Adam, Binary Crossentropy, Accuracy)\;
\end{itemize}

\textbf{Return:} Compiled CNN model\;

\end{algorithm*}

\subsection{Machine Unlearning Mechanism}


Machine unlearning refers to the concept where a model forgets or discards data used for training. This is particularly useful when certain data becomes irrelevant or when privacy issues arise (such as removing data associated with a specific individual). 

In the process of machine unlearning, the model is first trained on the entire dataset. For each sample, an influence score is computed based on the gradients of the model's predictions with respect to the loss function. This score measures how much a particular sample contributed to the model's learning. The influence score for sample \(i\) can be expressed as:

\[
\text{Influence Score for sample i} = \| \nabla \mathcal{L}(f(\mathbf{x}_i), y_i) \|
\]

where \( \nabla \mathcal{L} \) is the gradient of the loss function \( \mathcal{L} \) with respect to the model's prediction \( f(\mathbf{x}_i) \) for sample \( i \), and \( y_i \) is the true label.

Based on these influence scores, certain samples (such as the least influential ones) are removed from the training dataset. The rationale is that removing these samples will not significantly degrade the model's performance, but will help reduce the model's reliance on less useful or sensitive data. Formally, we can identify and discard the samples with the smallest influence scores:

\[
S_{\text{remove}} = \{ \mathbf{x}_i : \mathcal{I}(\mathbf{x}_i) \text{ is among the lowest in } S \}
\]

After removing the data points with the lowest influence scores, the model is retrained on the reduced dataset. This mimics a "forgetting" mechanism, as the model no longer has access to those removed data points, but retains the important features learned from the remaining data. Let \( S_{\text{new}} = S \setminus S_{\text{remove}} \) be the new training set. The model is retrained using the reduced dataset:

\[
\hat{f}_{\text{new}} = \arg \min_f \sum_{\mathbf{x}_i \in S_{\text{new}}} \mathcal{L}(f(\mathbf{x}_i), y_i)
\]

\begin{figure*}[!htb]
        \centering
		\includegraphics[scale=1.1]{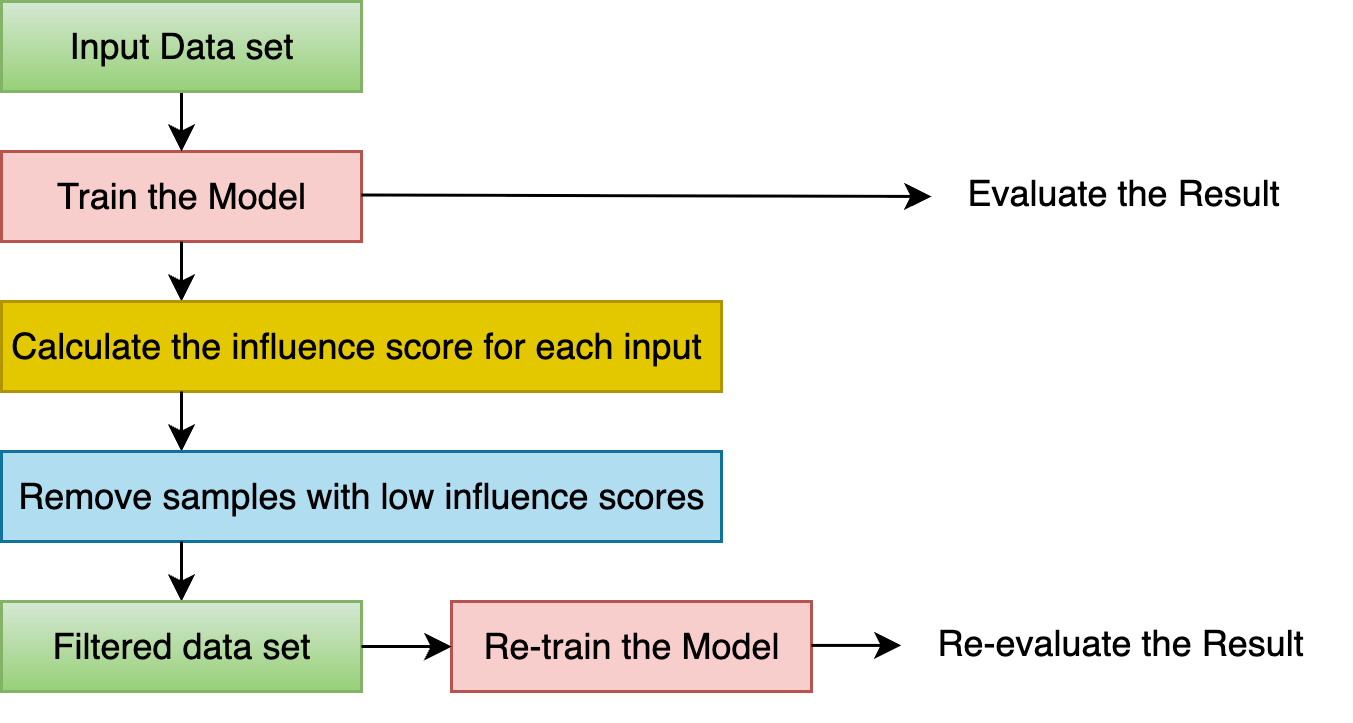}
		\caption{Machine Unlearning Framework} \label{frame_ul2}
\end{figure*}

Thus, the model \( \hat{f}_{\text{new}} \) is updated without the influence of the removed data points, effectively forgetting them while retaining the learned features from the remaining data. Figure \ref{frame_ul2} illustrates the overall machine unlearning framework, depicting key steps from influence score calculation to data removal and model retraining.

The influence score for each image in the dataset is calulated based on the gradients of the model’s predictions with respect to the input image, as explained below. The influence score indicates how much an individual image influences the model’s decision. The higher the influence score, the more sensitive the model's prediction is to changes in that image.

\subsubsection{Influence Score Calculation}
To quantify how much a sample contributes to learning, we define an influence score as follows.

Given a trained model $M$ with parameters $\theta$, the prediction for an input image $\mathbf{x}$ is 
\[
\hat{y} = f(\mathbf{x}; \theta).
\] 
To quantify the effect of an image on the model’s decision, we compute an \textit{influence score} using gradients. The procedure consists of the following steps:

\begin{enumerate}
    \item \textbf{Gradient Computation:}  
    For each image, compute the gradient of the prediction $\hat{y}$ with respect to the input $\mathbf{x}$:  
    \[
    \nabla_{\mathbf{x}} \hat{y} = \frac{\partial \hat{y}}{\partial \mathbf{x}}.
    \]  
    This captures pixel-level sensitivity.

    \item \textbf{Flattening:}  
    Each gradient tensor is flattened into a one-dimensional vector $\mathbf{g}$ for ease of handling.

    \item \textbf{Concatenation:}  
    Gradients from all layers, $\mathbf{g}_1, \dots, \mathbf{g}_n$, are concatenated into a single vector:  
    \[
    \mathbf{v} = \text{Concat}(\mathbf{g}_1, \dots, \mathbf{g}_n).
    \]

    \item \textbf{Norm Calculation:}  
    The L2 norm of $\mathbf{v}$ gives the influence score:  
    \[
    s = \|\mathbf{v}\|_2 = \sqrt{\sum_i v_i^2}.
    \]

    \item \textbf{Final Result:}  
    Repeating the above steps for the dataset $D$ yields an array of scores:  
    \[
    \mathbf{S} = [s_1, s_2, \dots, s_N].
    \]
\end{enumerate}

\begin{algorithm*}
\caption{Calculate Influence Scores}
\label{alg:influence}
\KwIn{Trained model $M$, DataFrame $D$ with image filenames and labels}
\KwOut{Array of influence scores $S$}

Initialize empty list $S \gets [~]$\;

\ForEach{row $r$ in $D$}{
    $p \gets$ filename from $r$\;
    $y \gets$ label from $r$ converted to float32\;
    
    $X \gets$ \texttt{PreprocessImageForCustomCNN}($p$)\;
    
    $G \gets$ \texttt{ComputeGradients}($M$, $X$, ExpandDims($y$))\;
    
    Initialize empty list $F$\;
    
    \ForEach{gradient tensor $g$ in $G$}{
        Append $\texttt{Reshape}(g, (-1))$ to $F$\;
    }
    
    $V \gets \texttt{Concat}(F)$\;
    
    $s \gets \|\ V \ \|_2$\;
    
    Append $s$ to $S$\;
}

\Return $S$\;
\end{algorithm*}
Algorithm~\ref{alg:influence} outlines the step-by-step procedure for computing influence scores. 
For each image in the dataset, the model gradients are computed, flattened, and concatenated across layers. 
The L2 norm of this concatenated gradient vector yields a single scalar influence score, which quantifies the sensitivity of the model’s prediction to that image. 
By iterating over the dataset, the algorithm produces an array of influence scores that highlights which images exert the strongest effect on the model’s decision-making. These influence scores enable targeted data removal to facilitate efficient and effective machine unlearning.



\section{Results and Discussion}
The model was implemented in Python version 3.10.14 using machine learning libraries such as \textit{Keras}, \textit{TensorFlow}, \textit{Scikit-learn}, \textit{Pandas}, \textit{NumPy}, and \textit{Matplotlib}. Training was performed with a batch size of 128 for 50 epochs, optimized with the \textit{Adam} optimizer (initial learning rate = 0.001) and \textit{categorical cross-entropy} as the loss function.
\subsection{Performance of the Proposed Model}
The customized CNN model was evaluated under three distinct scenarios: (i) training without augmentation, (ii) training with augmentation, and (iii) retraining with augmentation after machine unlearning (removal of 5\% training data). 

Eight augmentation techniques were applied to the training dataset, including rescaling, rotation, width and height shifting, shear transformation, zooming, and horizontal flipping. These parameters (Table \ref{tab:aug_para}) expanded data variability and helped mitigate overfitting. Their role in stabilizing training dynamics is further highlighted in the learning curve analysis (Fig. \ref{fig:learning_curves}).
\begin{table}[ht]
    \centering
    \caption{Data Augmentation Parameters Used for Model Training}
    \label{tab:aug_para}
    \begin{tabular}{|c|c|}
        \hline
        \textbf{Parameter} & \textbf{Value} \\
        \hline
        Rescale & \( \frac{1}{255} \) \\
        Rotation Range & 30 \\
        Width Shift Range & 0.2 \\
        Height Shift Range & 0.2 \\
        Shear Range & 0.2 \\
        Horizontal Flip & True \\
        Vertical Flip & True \\
        Fill Mode & Nearest \\
        \hline
    \end{tabular}
\end{table}

\begin{figure*}[!h]
    \centering
    \includegraphics[scale=0.6]{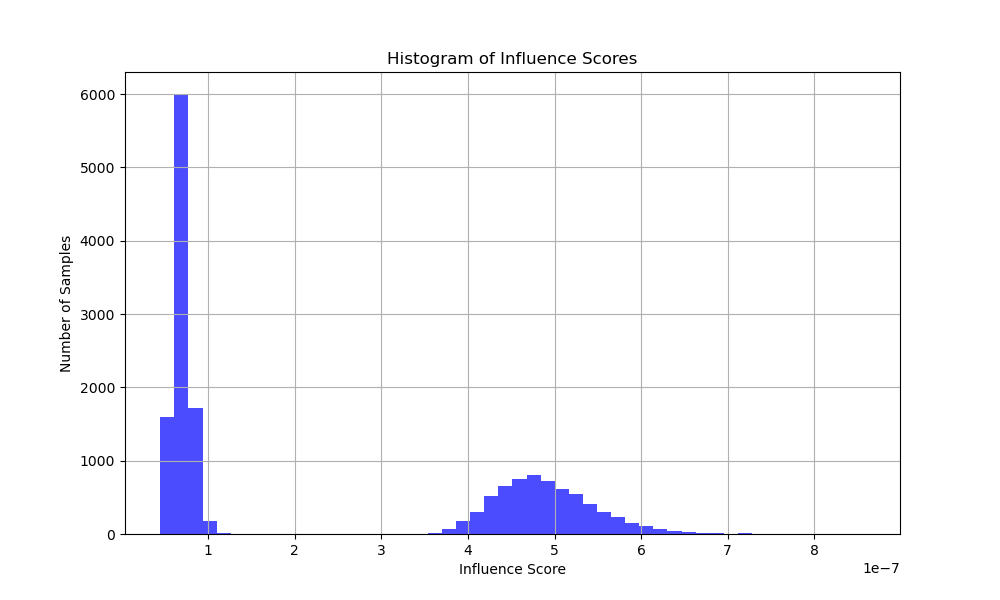}
    \caption{Distribution of influence scores for training samples (50 bins). }
    \label{Custom_influence}
\end{figure*}

To quantify the contribution of individual training samples, influence scores were calculated as described in Section~3.1.1. The histogram of scores (Fig. \ref{Custom_influence}) revealed that while most samples exerted moderate influence, a subset exhibited either very high or very low values. In the proposed unlearning strategy, the least influential 5\% of samples were excluded, since their removal was expected to reduce noise and redundancy without significantly degrading model performance. This formed the basis of the machine unlearning step, aimed at reducing reliance on less useful data. The resulting reduced dataset was then used to retrain the model, effectively forgetting the discarded samples while retaining the critical patterns learned from the remaining data.

\subsubsection{Learning curves}
\begin{figure*}[t]
  \centering
  
  \begin{subfigure}[b]{0.32\textwidth}
    \centering
    \includegraphics[width=\linewidth]{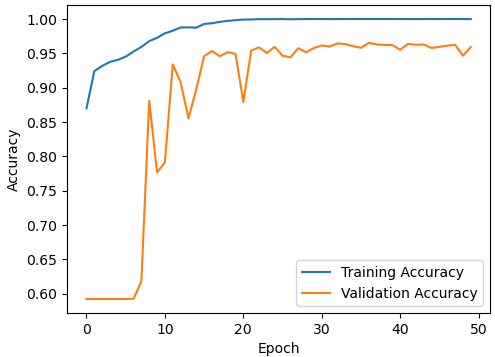}\\[4pt]
    \includegraphics[width=\linewidth]{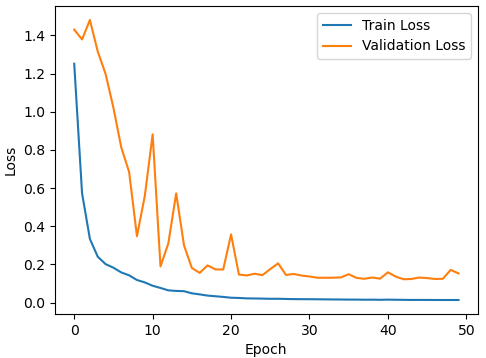}
    \caption{No Augmentation}
    \label{fig:no_aug}
  \end{subfigure}
  \hfill
  \begin{subfigure}[b]{0.32\textwidth}
    \centering
    \includegraphics[width=\linewidth]{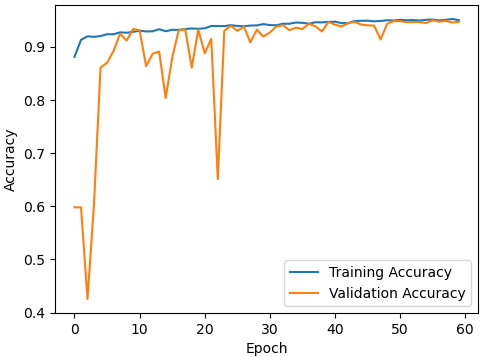}\\[4pt]
    \includegraphics[width=\linewidth]{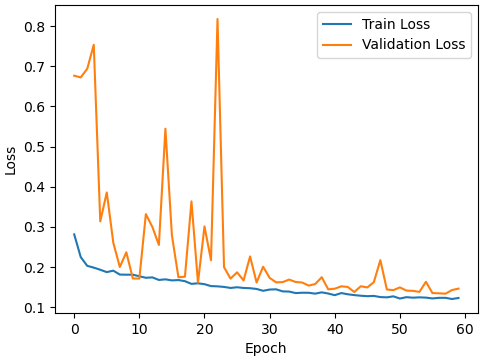}
    \caption{With Augmentation}
    \label{fig:aug}
  \end{subfigure}
  \hfill
  \begin{subfigure}[b]{0.32\textwidth}
    \centering
    \includegraphics[width=\linewidth]{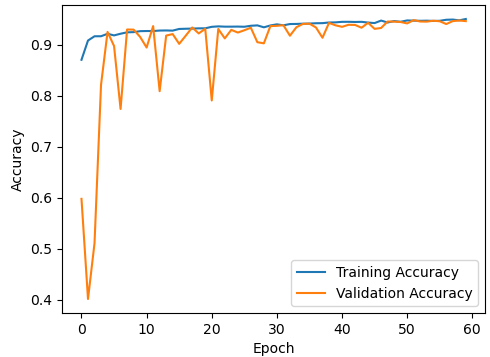}\\[4pt]
    \includegraphics[width=\linewidth]{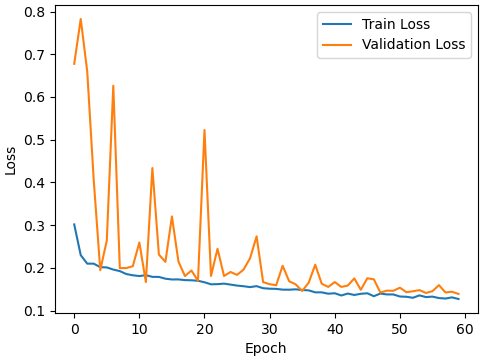}
    \caption{With Augmentation + MU (5\%)}
    \label{fig:aug_mu}
  \end{subfigure}

  \caption{Learning curves (accuracy and loss) for the CNN under three scenarios:
  (a) without augmentation, (b) with augmentation, and (c) with augmentation + machine unlearning (5\% data removal). 
  Each subfigure shows Accuracy (top) and Loss (bottom).}
  \label{fig:learning_curves}
\end{figure*}

The training and validation behavior of the CNN under different experimental settings is shown in Fig \ref{fig:learning_curves}. In the no augmentation scenario (Fig. \ref{fig:no_aug}), training accuracy increased rapidly and plateaued near 0.99, whereas validation accuracy fluctuated widely during the first 20 epochs before stabilizing around 0.95. The divergence between training and validation loss, along with the instability of validation loss, indicates overfitting and limited generalization despite high training accuracy.

With the introduction of data augmentation (Fig. \ref{fig:aug}), both training and validation curves exhibited improved stability. Validation accuracy consistently tracked training accuracy above 0.90, and validation loss displayed fewer spikes compared to the no-augmentation setting. These results confirm that augmentation enhanced generalization and convergence stability.

The augmentation with machine unlearning (5\%) configuration (Fig. \ref{fig:aug_mu}) produced the most stable training dynamics. Training and validation accuracies converged smoothly above 0.93, with both loss curves showing steady declines and minimal divergence. Importantly, validation fluctuations were further reduced compared to the augmentation-only scenario, suggesting that machine unlearning contributed to robust generalization, reduced overfitting, and improved reliability across epochs.

\subsubsection{Confusion Matrices and Classification Report}

\begin{figure*}[htbp]
    \centering
    
    \begin{subfigure}[b]{0.25\linewidth}
        \centering
        \includegraphics[width=\linewidth]{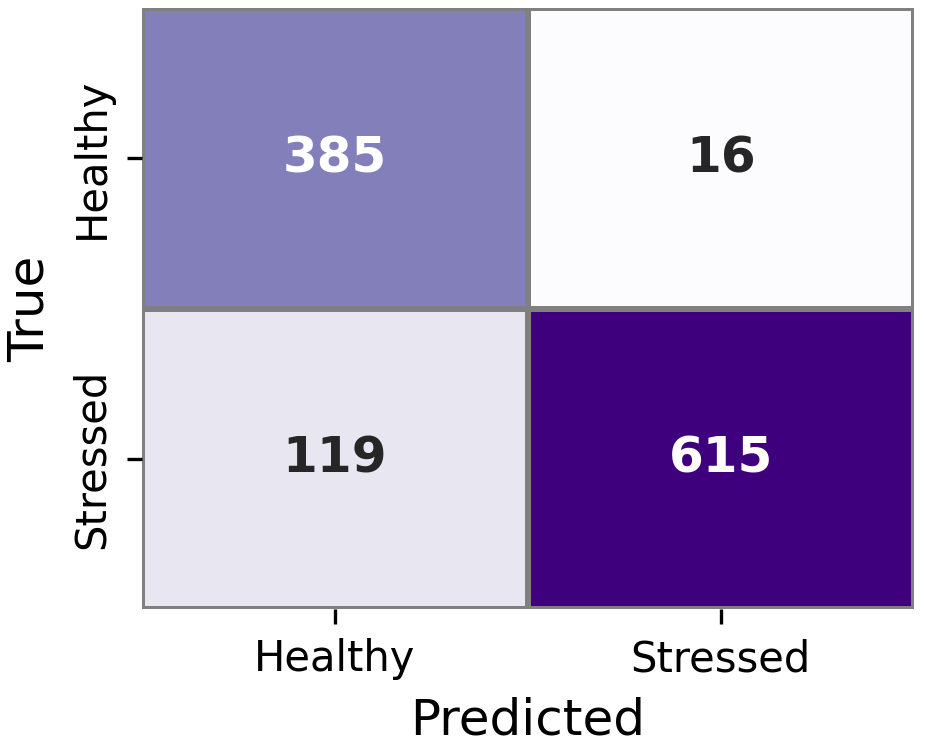}
        \caption{No Augmentation}
        \label{fig:conf1}
    \end{subfigure}
    \hfill
    \begin{subfigure}[b]{0.25\linewidth}
        \centering
        \includegraphics[width=\linewidth]{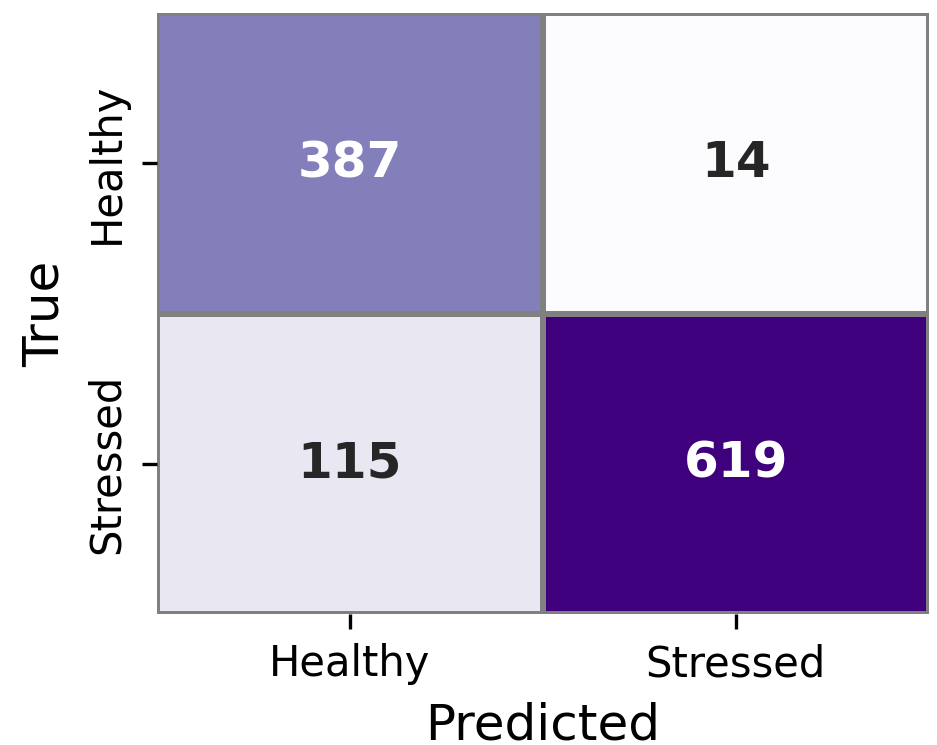}
        \caption{With Augmentation}
        \label{fig:conf2}
    \end{subfigure}
    \hfill
    \begin{subfigure}[b]{0.25\linewidth}
        \centering
        \includegraphics[width=\linewidth]{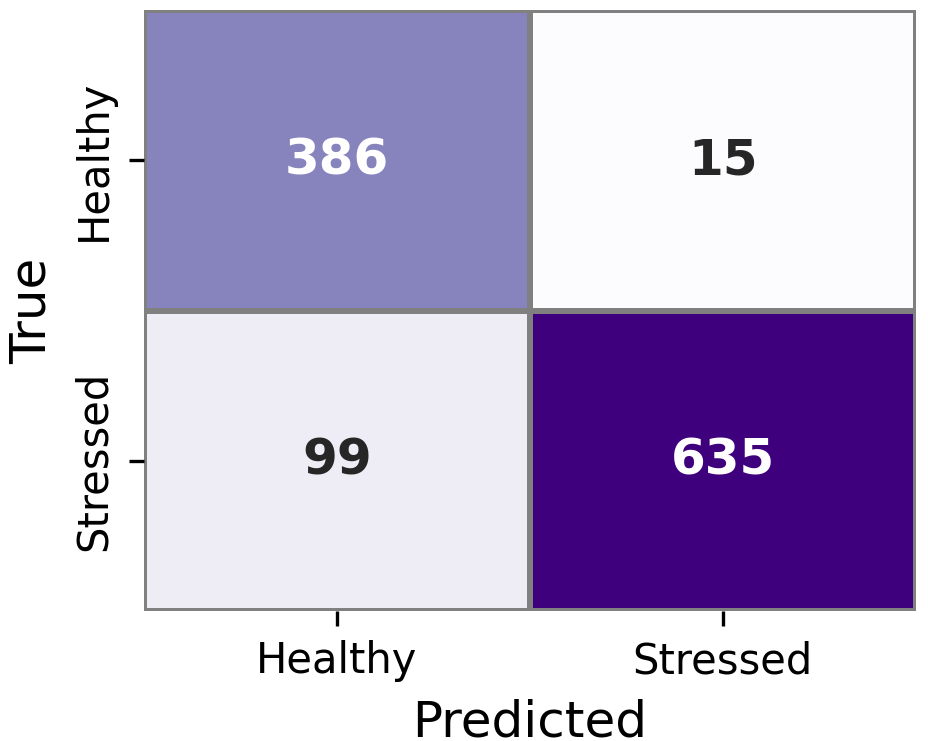}
        \caption{With Augmentation+Machine Unlearning}
        \label{fig:conf3}
    \end{subfigure}
    
    \caption{Confusion matrices for the three scenario.}
    \label{fig:con_mat}
\end{figure*}

The quantitative evaluation results are summarized in Table \ref{tab:scenarios_comparison}, with confusion matrices shown in Fig. \ref{fig:con_mat}. 

\textbf{Scenario-wise classification outcomes}
In the no augmentation setting, the framework achieved an overall accuracy of 88.1\%. The stressed class exhibited high precision (0.97) but comparatively lower recall (0.84), leading to mis-classification of stressed samples as healthy. The confusion matrix (Fig. \ref{fig:conf1}) confirms this, with 119 stressed samples incorrectly labeled as healthy, highlighting the model’s tendency to under-detect stress conditions despite the healthy class achieving a strong recall of 0.96.

The augmentation-only scenario improved both stability and generalization. Accuracy rose slightly to 88.6\%, with stressed precision maintained at 0.98 and recall remaining at 0.84. The confusion matrix (Fig. \ref{fig:conf2} illustrates a modest reduction in misclassifications, with 115 stressed samples misclassified as healthy and only 14 false positives among healthy samples. This indicates that augmentation reduced the imbalance and improved class separability.

The augmentation with machine unlearning (Aug + MU, 5\%) scenario yielded the best overall results. Accuracy peaked at 90.0\%, with stressed recall improving to 0.87 and F1-score to 0.92, while the healthy class achieved precision of 0.80 and recall of 0.96. As depicted in the confusion matrix (Fig. \ref{fig:conf3}), false negatives for stressed samples decreased to 99, the lowest across all scenarios, with only 15 false positives for the healthy class. This demonstrates that machine unlearning reduced error rates, and minimized false negatives compared to the other two settings.

\textbf{Comparative insights}
Taken together, the results from Table \ref{tab:scenarios_comparison} and Fig. \ref{fig:con_mat} highlight that stressed class predictions benefited the most from augmentation and unlearning, as recall consistently improved (0.84 → 0.87). Meanwhile, the healthy class maintained strong recall across all scenarios but exhibited notable gains in precision under the Aug + MU configuration. These improvements emphasize that the integration of data augmentation with machine unlearning achieves the best trade-off between precision and recall, yielding a more generalizable and balanced framework for classification.

\begin{table*}[t]
\centering
\caption{Comparative performance summary of the proposed framework in three scenarios. Bold values indicate the best within each column.}
\label{tab:scenarios_comparison}
\begin{tabular}{lccccccc p{2.8cm}}
\toprule
\multirow{2}{*}{Scenario} & \multicolumn{3}{c}{Stressed} & \multicolumn{3}{c}{Healthy} & \multirow{2}{*}{Accuracy} & \multirow{2}{*}{Key Notes} \\
\cmidrule(lr){2-4} \cmidrule(lr){5-7}
 & Precision & Recall & F1-score & Precision & Recall & F1-score & & \\
\midrule
No Augmentation              
  & 0.97 & 0.84 & 0.90 
  & 0.76 & \textbf{0.96}\textsuperscript{*} & 0.85 
  & 0.881 
  & \makecell[l]{Overfits, \\ unstable validation} \\

With Augmentation            
  & \textbf{0.98}\textsuperscript{*} & 0.84 & 0.91 
  & 0.77 & \textbf{0.97}\textsuperscript{*} & 0.86 
  & 0.886 
  & \makecell[l]{Better generalization, \\ stable curves} \\

With Aug + MU (5\%)          
  & \textbf{0.98}\textsuperscript{*} & \textbf{0.87} & \textbf{0.92} 
  & \textbf{0.80} & 0.96 & \textbf{0.87} 
  & \textbf{0.900} 
  & \makecell[l]{Best balance, \\robust generalization, \\ fewer false negatives} \\
\bottomrule
\end{tabular}
\end{table*}

\subsection{Comparison of Performance}
\begin{table*}[t]
\centering
\caption{Comparative performance evaluation of various existing works}
\label{tab:models_comparison}
\resizebox{\linewidth}{!}{  
\begin{tabular}{lcccccccc}
\toprule
\multirow{2}{*}{Model} & \multicolumn{2}{c}{Stressed} & \multicolumn{2}{c}{Healthy} & \multirow{2}{*}{\parbox{1.5cm}{F1-Score \\ (Stressed)}} & \multirow{2}{*}{\parbox{1.5cm}{F1-Score \\ (Healthy)}} & \multirow{2}{*}{\parbox{2cm}{Test \\ Accuracy}} & \multirow{2}{*}{\parbox{2cm}{Trainable \\ Parameters}} \\
\cmidrule(lr){2-3} \cmidrule(lr){4-5}
 & Precision & Recall & Precision & Recall &  &  &  &  \\
\midrule
MobileNet based Pipeline \cite{patra2024explainable} & 0.978 & 0.845 & 0.773 & 0.965 & 0.906 & 0.862 & 0.887 & 3.5 Million \\
DenseNet121 based Pipeline \cite{patra2024explainable} & 0.967 & 0.887 & 0.820 & 0.945 & 0.925 & 0.880 & 0.907 & 7.09 Million \\
ViT-TL \cite{patra2025explainable} & 0.968 & 0.901 & 0.839 & 0.945 & 0.934 & 0.890 & 0.916 & 14M \\
Proposed Framework (Aug + MU) \textsuperscript{*} & 0.980 & 0.870 & 0.800 & 0.960 & 0.922 & 0.874 & 0.900 & 0.231M \\
\bottomrule
\end{tabular}
}
\end{table*}

The comparative performance of the proposed CNN framework against existing state-of-the-art models is summarized in Table \ref{tab:models_comparison}. Prior pipelines such as MobileNet \cite{patra2024explainable} and DenseNet121 \cite{patra2024explainable} achieved respectable accuracies of 88.7\% and 90.7\%, respectively, while the ViT-TL model \cite{patra2025explainable} reached the highest reported accuracy of 91.6\%. These transformer-based and deep CNN architectures, however, come with significantly larger parameter counts (ranging from 3.5M to 12M), which demand higher computational resources and longer training times.  

In contrast, the proposed framework attains a competitive accuracy of 90.0\% with stressed and healthy F1-scores of 0.922 and 0.874, respectively. Notably, this is achieved with only 0.231M trainable parameters, representing a 15 to 60 fold reduction in model complexity compared to existing pipelines. The lightweight design translates to reduced memory footprint, faster inference, and improved deployability on edge devices or low-resource settings, where computational constraints often limit the use of heavier architectures.  

Another critical insight lies in class-specific performance. While MobileNet and DenseNet pipelines tend to favor healthy class recall, and ViT-TL maximizes stressed class recall, our proposed model achieves a more balanced trade-off between precision and recall across both classes. This is particularly important for stress detection tasks, where false negatives (misclassifying stressed plants as healthy) carry greater risk for agricultural monitoring and decision-making. The machine unlearning strategy further reduced such errors, resulting in the lowest stressed-class false negative rate (13.5\%) among the evaluated models.  

Taken together, these findings highlight that the proposed framework delivers state-of-the-art performance with minimal computational overhead, making it well-suited for real-world applications where resource efficiency and balanced predictive power are equally critical. By bridging the gap between accuracy and efficiency, the framework offers a practical alternative to heavy-weight architectures, without compromising generalization.

\section{Conclusion}

This study proposed a lightweight hybrid CNN integrated with gradient-guided machine unlearning for drought stress identification in potato crops. The framework achieved 90.0\% accuracy with only 0.231M parameters, offering state-of-the-art performance with at least 15-fold fewer parameters than existing models. Unlike heavier CNN or transformer-based pipelines, the proposed model delivered a balanced trade-off between precision and recall, while reducing false negatives in stressed plants—a critical requirement for agricultural monitoring. Its compact design makes it highly suitable for real-time deployment in UAVs, edge devices, and other resource-constrained settings. Future work will emphasize expanding to multimodal imagery, improving interpretability, and validating performance across diverse crops and field conditions.


\bibliographystyle{unsrt}
\bibliography{refe}

\end{document}